\documentclass[11pt,letterpaper]{mystyle}
\usepackage{fancyhdr}
\usepackage{float}

\usepackage[utf8]{inputenc} 
\usepackage[T1]{fontenc}
\usepackage[numbers]{natbib}
\usepackage{adjustbox}
\usepackage{breakurl}    
\usepackage{hyperref}
\usepackage[edges]{forest}
\usepackage{subcaption}
\usepackage{soul}
\usepackage{multirow}
\usepackage[utf8]{inputenc} 
\usepackage{booktabs}       
\usepackage{amsfonts}       
\usepackage{nicefrac}      
\usepackage{microtype}     
\usepackage{xcolor}        
\usepackage{colortbl}
\usepackage{enumitem}
\usepackage{amssymb}
\usepackage{wrapfig}
\usepackage{courier}
\usepackage{bxcoloremoji}
\usepackage{CJKutf8}
\usepackage{tcolorbox}
\usepackage{array}
\usepackage{tabularx}

\newcommand{\github}{\raisebox{-1.5pt}{\includegraphics[height=1.05em]{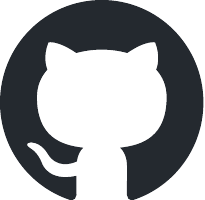}}}

\fancypagestyle{headstyle}{
    \fancyhead[L]{
        \includegraphics[width=80pt]{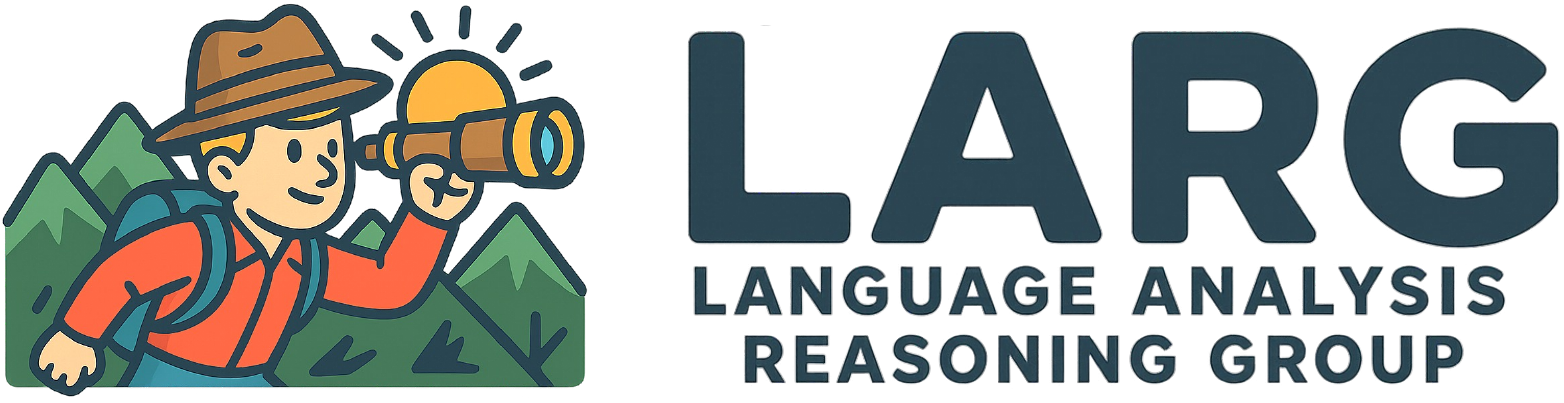}
    }
    \fancyhead[C]{}
    \fancyhead[R]{}

}

\definecolor{hidden-red}{RGB}{205, 44, 36}
\definecolor{hidden-blue}{RGB}{194,232,247}
\definecolor{hidden-orange}{RGB}{243,202,120}
\definecolor{hidden-green}{RGB}{34,139,34}
\definecolor{hidden-pink}{RGB}{255,245,247}
\definecolor{hidden-black}{RGB}{20,68,106}
\definecolor{purple}{RGB}{144,153,196}
\definecolor{yellow}{RGB}{255,228,123}
\definecolor{hidden-yellow}{RGB}{255,248,203}
\definecolor{tkcolor}{RGB}{224,223,255}
\definecolor{darkblue}{rgb}{0, 0.40, 0.75}

\hypersetup{colorlinks=true, citecolor=darkblue, linkcolor=darkblue, urlcolor=darkblue}

\newtcolorbox{socialpost}{
  colback=black!5!white,
  colframe=black!75!black,
  sharp corners,
  boxrule=0.5pt,
  fontupper=\sffamily\small, 
  left=3mm,
  right=3mm,
  top=3mm,
  bottom=3mm,
  boxsep=1mm
}

\tcbset{
  aibox/.style={
    width=\linewidth,
    top=8pt,
    bottom=4pt,
    colback=blue!6!white,
    colframe=black,
    colbacktitle=black,
    enhanced,
    center,
    attach boxed title to top left={yshift=-0.1in,xshift=0.15in},
    boxed title style={boxrule=0pt,colframe=white,},
  }
}

\newtcolorbox{AIbox}[2][]{aibox,title=#2,#1}

\tcbset{
  takeawaybox/.style={
    width=\linewidth,
    top=8pt,
    bottom=4pt,
    colback=hidden-yellow,
    colframe=black,
    colbacktitle=black,
    enhanced,
    center,
    attach boxed title to top left={yshift=-0.1in,xshift=0.15in},
    boxed title style={boxrule=0pt,colframe=white,},
  }
}

\newtcolorbox{TakeawayBox}[2][]{takeawaybox,title=#2,#1}

\newtcolorbox{promptbox}[2][]{
    width=\columnwidth,
    colback = gray!8, 
    colframe = gray!8, 
    boxsep=0pt,left=5pt,right=5pt,top=5pt,bottom=5pt,
    fontupper=\linespread{0.9}\selectfont,
    title=#2,#1
}

\title{Beware of Reasoning Overconfidence: Pitfalls in the Reasoning Process for Multi-solution Tasks}

\author{
  Jiannan Guan$^{1*}$ \quad 
  Qiguang Chen$^{1*}$ \quad 
  Libo Qin$^{2, \coloremojicode{2709}}$ \quad 
  Dengyun Peng$^{1}$ \quad 
  Jinhao Liu$^{1}$ \quad 
  Liangyu Huo$^{3}$ \quad 
  Jian Xie$^{3}$ \quad
  Wanxiang Che$^{1, \coloremojicode{2709}}$ \vspace{-5pt}\\
\normalfont{
$^1$ Research Center for Social Computing and Interactive Robotics, Harbin Institute of Technology,\vspace{-5pt}\\
$^2$ School of Computer Science and Engineering, Central South University,\vspace{-5pt}\\
$^3$ Du Xiaoman (Beijing) Science Technology Co., Ltd. \\
}}

\begin{document}

\begin{abstract}
  \vspace{5mm}
  \textbf{\large Abstract:}
  \vspace{2mm}

  Large Language Models (LLMs) excel in reasoning tasks requiring a single correct answer, but they perform poorly in multi-solution tasks that require generating comprehensive and diverse answers. We attribute this limitation to \textbf{reasoning overconfidence}: a tendency to express undue certainty in an incomplete solution set. To examine the effect, we introduce \textit{MuSoBench}, a benchmark of multi-solution problems. Experiments show that the conventional short chain-of-thought (Short-CoT) prompting paradigm exhibits pronounced overconfidence, whereas the emerging long chain-of-thought (Long-CoT) approach mitigates it through iterative exploration and self-reflection. We further characterise observable behaviours and influential factors. To probe the underlying cause, we propose the \textbf{cognitive-rigidity hypothesis}, which posits that overconfidence arises when the reasoning process prematurely converges on a narrow set of thought paths. An attention-entropy analysis offers preliminary support for this view. These findings provide tools for assessing the completeness of LLM reasoning and highlight the need to move evaluation beyond single-answer accuracy toward comprehensive exploration.
  \vspace{5mm}

  $^{*}$ \textit{Equal Contribution}

  $^{\coloremojicode{2709}}$ \textit{Corresponding Author}

  \vspace{5mm}

  \coloremojicode{1F4C5} \textbf{Date}: Oct. 01, 2025


  \github{} \textbf{Code Repository}: \href{https://github.com/jubgjf/reasoning-overconfidence}{https://github.com/jubgjf/reasoning-overconfidence}



  \coloremojicode{1F4E7} \textbf{Contact}: \href{mailto:jnguan@ir.hit.edu.cn}{jnguan@ir.hit.edu.cn}, \href{mailto:qgchen@ir.hit.edu.cn}{qgchen@ir.hit.edu.cn}, \href{mailto:lbqin@csu.edu.cn}{lbqin@csu.edu.cn}, \href{mailto:car@ir.hit.edu.cn}{car@ir.hit.edu.cn}

\end{abstract}
\maketitle

\vspace{3mm}
\pagestyle{headstyle}
\thispagestyle{empty}

\begin{CJK}{UTF8}{gbsn}

\section{Introduction}

\begin{figure}[t]
    \centering
    \includegraphics[width=0.98\linewidth]{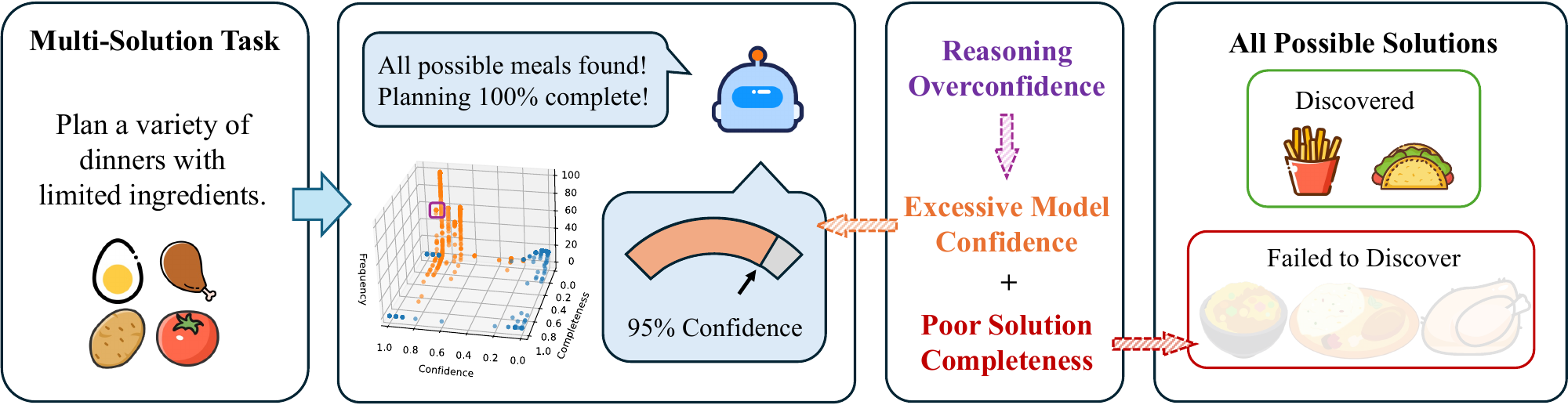}
    \caption{
        On multi-solution tasks, the model suffers from \textbf{reasoning overconfidence}, exhibiting excessively high confidence while exploring only a few reasoning paths.
        This leads to a poor completeness score for the final task.
    }
    \label{fig:figure-1}
\end{figure}

Recently, Large Language Models (LLMs) have shown strong performance on tasks requiring multiple correct answers~\cite{achiam2023gpt,yang2025qwen3,zhuang2023through,qin2024large}.
As illustrated in Figure~\ref{fig:figure-1}, consider planning every possible dinner from a fixed set of ingredients: success lies in listing the full menu, not a single dish.
We call such problems multi-solution reasoning tasks, whose goal is completeness and diversity.
Yet advanced methods like Chain-of-Thought (CoT)~\cite{wei2022chain}, designed for one reasoning path, often stop early.
When asked to list all answers, LLMs usually produce a few options and then assert confidently that no others exist.
As Figure~\ref{fig:figure-1} \& \ref{fig:recall-vs-confidence-3d} depict, this overconfidence significantly reduces possible solution exploration, exposing a mismatch between stated confidence and actual coverage.
To systematically analyze this failure mode on multi-solution tasks, we introduce the concept of \textbf{Reasoning Overconfidence}:
A model’s subjective confidence in its solution set significantly exceeds its actual ability to recover the full set of correct answers.

Prior work has examined LLM performance on multi-solution tasks.
Some work focuses on reasoning under structured constraints.
For example, the 24-point game requires enumerating all valid arithmetic expressions~\cite{yao2023tree}, while benchmarks such as CalibratedMath use problems with multiple correct answers to assess uncertainty calibration~\cite{lin2022teaching}.
Others center on open-ended generation.
Creative tasks such as story generation evaluate producing diverse content from an effectively unbounded solution space~\cite{xu2024echoes,wang2024guiding}.
However, the datasets used in existing work share a fundamental limitation:
Their solution spaces are either tightly constrained or nearly unbounded, which hinders reliable estimation of completeness.

To enable more precise empirical study, we developed the \textbf{Multi-solution Benchmark (MuSoBench)}, a task suite designed to evaluate reasoning overconfidence under controlled solution spaces.
When applied to MuSoBench, the conventional short chain-of-thought (Short-CoT) approach exhibits persistent overconfidence: as shown in Figure \ref{fig:recall-vs-confidence-3d}, its outputs cluster in the high-confidence, low-recall region, providing direct empirical evidence of this behavior. Behavioral analysis indicates that Short-CoT performs a shallow search, seldom revising its initial reasoning path, which largely explains its inflated confidence.
By contrast, the Long-CoT paradigm, which promotes iterative exploration and self-reflection~\cite{chen2025aware,chen2025towards,zeng2024scaling,li2025system}, substantially improves both recall and precision, thereby reducing reasoning overconfidence.
Finally, analysis of internal activations supports the cognitive rigidity hypothesis, which attributes overconfidence to premature convergence on a narrow set of reasoning paths.

The main contributions are summarized as follows:
\begin{itemize}[leftmargin=16pt, itemsep=0pt, topsep=0pt]
    \item We first introduce the reasoning overconfidence concept as a critical failure mode of LLMs on multi-solution tasks and present MuSoBench, a new benchmark that documents this phenomenon through evidence on solution diversity, stability, and calibration.
    \item We analyze factors influencing overconfidence and its mitigation. Our results show that the extent of overconfidence is governed chiefly by the length of the reasoning trace, the presence of reflective steps, and the breadth of exploration, thereby linking the phenomenon to both reasoning paradigms and task properties.
    \item We advance the cognitive-rigidity hypothesis to explain this behavior, examining internal model states that give rise to overconfidence and offering a fresh perspective on the fundamental multi-solution reasoning.
\end{itemize}

\section{Problem Formulation \& Benchmark}

\subsection{Multi-solution Tasks}
Multi-solution tasks require a model to enumerate all valid answers to a single problem rather than to return just one.
Formally, such a task is characterised by a dataset $\mathcal{T}=\{(x_i, \hat{\mathcal{Y}}_i) \ | \ | \hat{\mathcal{Y}}_i | \ge 1\}_{i=1}^N$, where $x_i$ denotes the $i$-th problem instance and $\hat{\mathcal{Y}}_i$ is the corresponding ground-truth solution set.
Given $x_i$, a model $\mathcal{M}$ outputs its own solution set $\mathcal{Y}_i = \mathcal{M}(x_i)$.
The ideal outcome is that $\mathcal{Y}_i$ matches $\hat{\mathcal{Y}}_i$ exactly, covering every valid solution and omitting none.

\begin{figure}[t]
    \centering
    \includegraphics[width=0.98\linewidth]{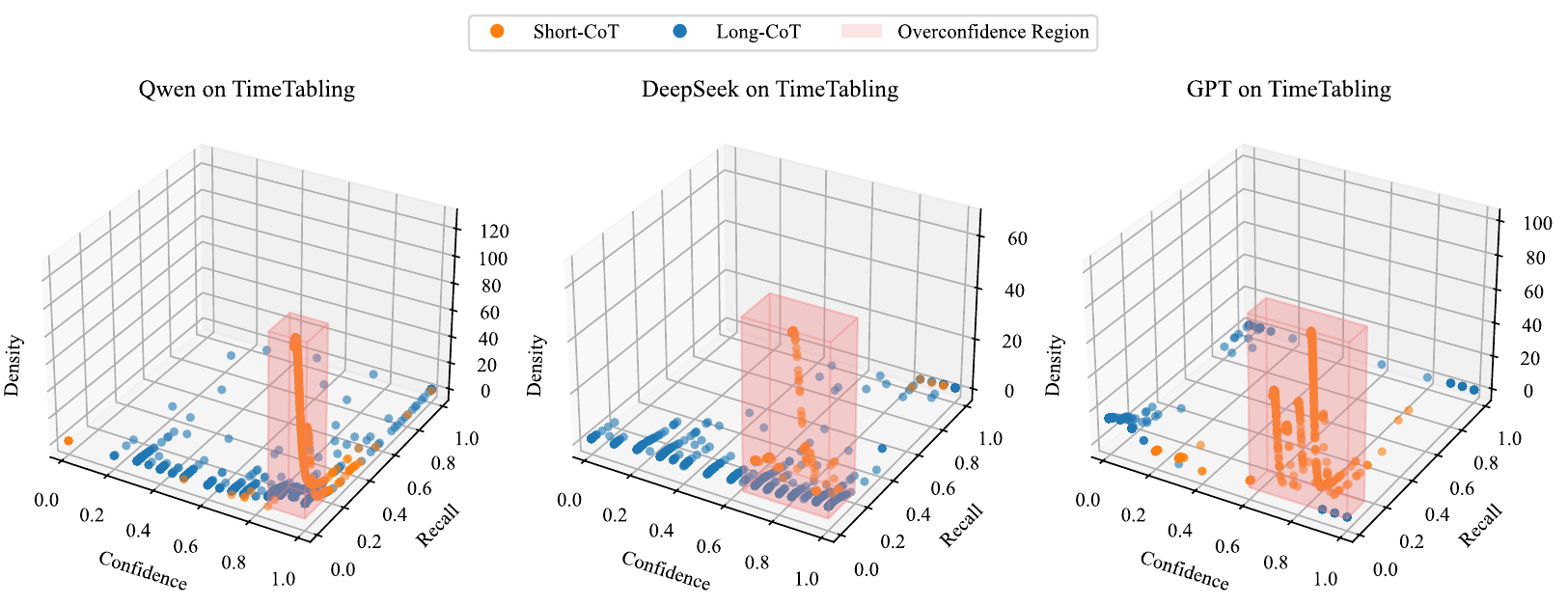}
    \caption{
        Distribution plots of recall vs. confidence on the TimeTabling dataset.
        The plots clearly show Short-CoT results clustering in the low-recall, high-confidence corner (red).
        For SubsetSum results, see Figure~\ref{fig:recall-vs-confidence-3d-subsetsum} in Appendix.
    }
    \label{fig:recall-vs-confidence-3d}
\end{figure}

\subsection{Reasoning Overconfidence}

\begin{figure*}[t]
    \centering
    \includegraphics[width=0.98\linewidth]{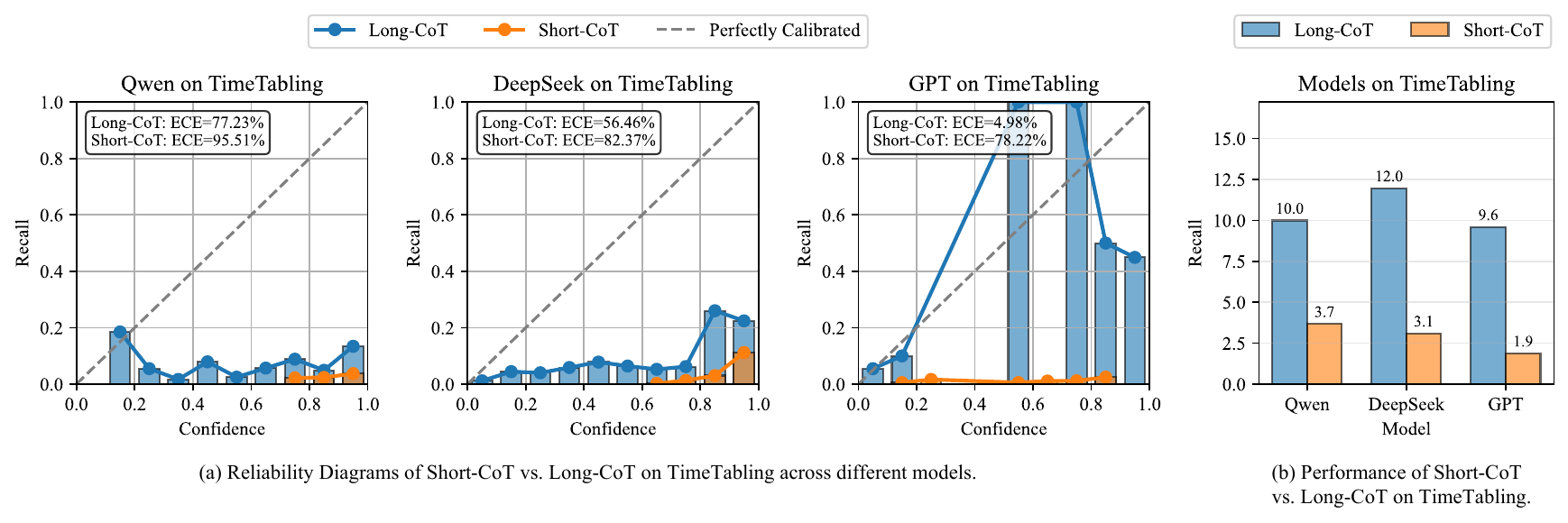}
    \caption{
        Calibration and performance of Short-CoT vs. Long-CoT on TimeTabling dataset.
        As shown in (a), the diagonal line represents perfect calibration.
        Long-CoT models (blue) are better calibrated than Short-CoT models (orange).
        As shown in (b), Long-CoT models achieve significantly higher recall than Short-CoT models.
        For SubsetSum results, see Figure~\ref{fig:reliability-diagrams-and-short-cot-long-cot-recall-subsetsum} in Appendix.
    }
    \label{fig:reliability-diagrams-and-short-cot-long-cot-recall}
\end{figure*}

\textbf{R}easoning \textbf{O}ver\textbf{C}onfidence (ROC) refers to a model’s tendency to report confidence levels that exceed its actual performance. In multi-solution tasks, this mismatch often drives the model to return only a subset of correct answers and to stop searching prematurely.
To characterise the phenomenon quantitatively, we use the \emph{Expected Calibration Error} (ECE), which compares expected and realised performance. Let $\mathcal{C}(\mathcal{Y}_i \mid x_i, \mathcal{M})$ denote the confidence that model $\mathcal{M}$ assigns to its proposed solution set $\mathcal{Y}_i$.
A model is \emph{reasoning-overconfident} when its reported confidence exceeds its observed performance $\mathrm{Perf}(\cdot)$:
\begin{equation}
    \mathcal{C}(\mathcal{Y}_i|x_i,\mathcal{M}) > \text{Perf}(\mathcal{Y}_i, \hat{\mathcal{Y}}_i)
\end{equation}
For tasks with multiple valid answers, the relevant performance dimension is the \textbf{completeness} of the returned set. Because our focus is the model’s failure to enumerate \emph{all} correct solutions, not the precision of each individual answer, we measure performance with \textbf{Recall}. Metrics that combine precision and recall (e.g., F1-score) would blur the exploration shortfall we seek to isolate; hence, we instantiate $\mathrm{Perf}(\cdot)$ as Recall throughout this work.

\subsection{MuSoBench Construction}

To systematically and controllably evaluate models on multi-solution tasks, we introduce the \textbf{Multi-Solution Benchmark (MuSoBench)}, comprising two subtasks: \emph{TimeTabling} and \emph{SubsetSum}.
The \emph{TimeTabling} subtask is to construct conflict-free course schedules subject to constraints on course overlap, instructor availability, and classroom capacity.
The \emph{SubsetSum} subtask requires enumerating all subsets of a given integer set summing to target number.

Problem complexity is measured by the size of each instance’s solution space.
The TimeTabling corpus spans ten complexity levels and the SubsetSum corpus seven, with 100 instances per level.
For every instance, we algorithmically enumerated all feasible solutions and manually verified them to ensure correctness and completeness.
A detailed description of the dataset construction procedure, together with illustrative examples, is provided in Appendix~\ref{appsec:benchmark-construction}.

To quantitatively assess model behavior in multi-solution scenarios, we utilize the following metrics:

\begin{itemize}[leftmargin=16pt, itemsep=0pt, topsep=0pt]
    \item \textbf{Model Performance Metric:}
          (1) \textit{Precision} ($\uparrow$): Proportion of generated answers that are correct.
          (2) \textit{Recall} ($\uparrow$): Proportion of ground-truth answers the model recovers. This is the primary metric for multi-solution tasks.
    \item \textbf{Overconfidence Metric:}
          \textit{Expected Calibration Error (ECE)} ($\downarrow$): Average gap between reported confidence and realized performance (precision or recall). Lower ECE indicates better calibration.
    \item \textbf{Model Behavior Metric:}
          (1) \textit{Correct Solution Retention Rate (CSR)} ($\uparrow$): Capability to maintain previously correct solutions.
          (2) \textit{Error Solution Correction Rate (ESC)} ($\uparrow$): Capability to correct earlier error solutions.
          (3) \textit{New Solution Discovery Rate (NSD)} ($\uparrow$): Capability to discover additional correct solutions.
\end{itemize}

All detailed formulas are in Appendix~\ref{appsec:evaluation-metrics}.

\begin{figure*}[t]
    \centering
    \includegraphics[width=0.90\linewidth]{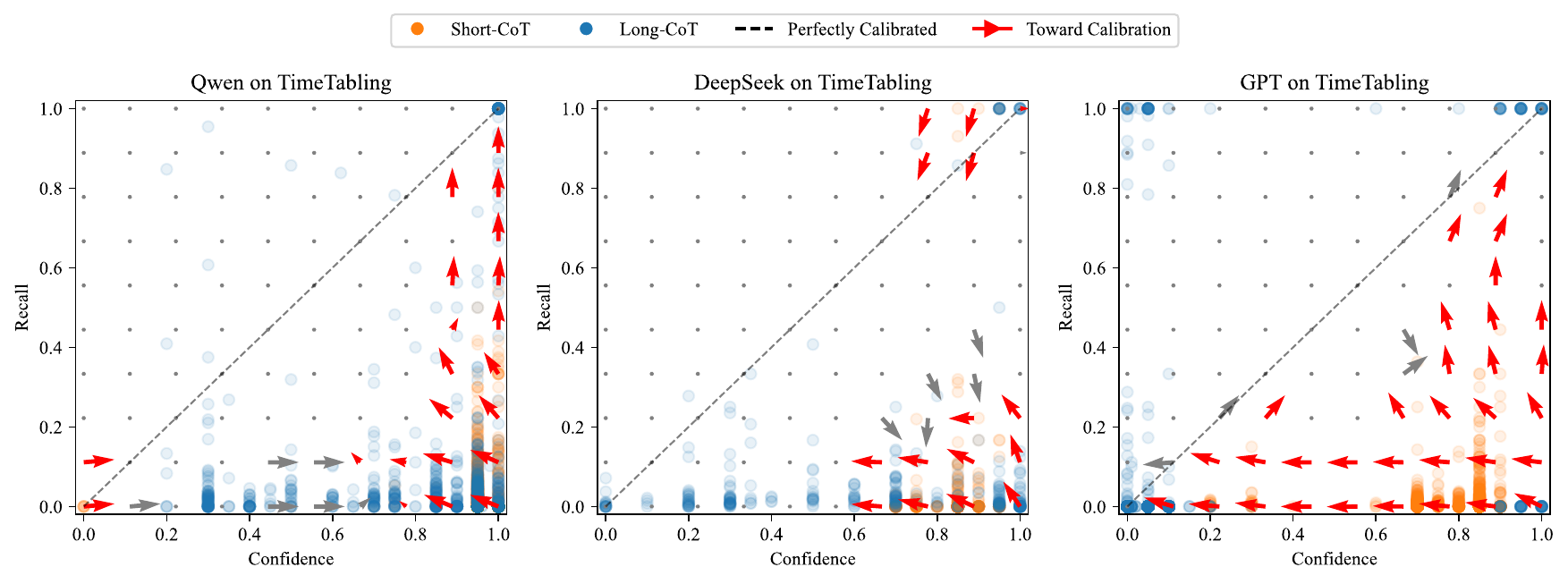}
    \caption{
        The arrows indicate the movement of model confidence and performance from Short-CoT to Long-CoT.
        The results show that adopting Long-CoT causes most data points to shift toward the diagonal, indicating improved calibration (red).
        Results for SubsetSum are shown in Figure~\ref{fig:recall-vs-confidence-movement-subsetsum} in Appendix.
    }
    \label{fig:recall-vs-confidence-movement}
\end{figure*}

\section{Experimental Setup}

Our experiments are conducted on the Qwen, DeepSeek, and GPT series of models.
We mainly compare the following two CoT reasoning paradigms:
\begin{itemize}[leftmargin=16pt, itemsep=0pt, topsep=0pt]
    \item \textbf{Short-CoT:} Appends zero-CoT prompt~\citep{kojima2022large} to a base instruction to elicit CoT from instruction-tuned models, including Qwen3-8B (non-thinking mode)~\cite{yang2025qwen3}, DeepSeek-V3~\cite{liu2024deepseek}, and gpt-4o-mini~\cite{achiam2023gpt}.
    \item \textbf{Long-CoT:} Uses models trained for extended, iterative reasoning and reflection; we use Qwen3-8B (thinking mode)~\cite{yang2025qwen3}, DeepSeek-R1~\cite{guo2025deepseek}, and o4-mini~\cite{jaech2024openai}.
\end{itemize}
We contend this is a valid comparison of the entire paradigm as deployed in practice, rather than comparing different prompts or training methods.

To estimate LLM confidence, we use a verbal-elicitation strategy~\cite{xiong2023can}. This approach applies to both open-source and API-only models and is supported by evidence that verbalized confidence closely tracks internal probabilities~\cite{kumar2024confidence}. In our setup, the model first generates a full answer and then reports its confidence on a \(0\)–\(100\) scale. The prompts are given in Appendix~\ref{appsec:confidence-elicitation}.

\section{Analysis of Reasoning Overconfidence}

\begin{table*}[t]
    \centering
    \small
    \begin{tabular}{cccccccc}
        \toprule
        \multirow{2}[1]{*}{Model}    & \multirow{2}[1]{*}{Paradigm} & \multicolumn{3}{c}{TimeTabling} & \multicolumn{3}{c}{SubsetSum}                                                                                         \\
        \cmidrule(lr){3-5} \cmidrule(lr){6-8}

                                     &                              & CSR (\%) $\uparrow$             & ESC (\%) $\uparrow$           & NSD (\%) $\uparrow$ & CSR (\%) $\uparrow$ & ESC (\%) $\uparrow$ & NSD (\%) $\uparrow$ \\
        \midrule
        \multirow{2}[0]{*}{Qwen}     & Short-CoT                    & \textbf{43.15}                  & 65.22                         & 0.19                & 14.89               & 15.63               & 0.20                \\
                                     & Long-CoT                     & 53.13                           & \textbf{81.83}                & \textbf{2.50}       & \textbf{75.65}      & \textbf{93.93}      & \textbf{0.49}       \\
        \midrule
        \multirow{2}[0]{*}{DeepSeek} & Short-CoT                    & 6.77                            & \textbf{98.86}                & 0.01                & \textbf{70.44}      & 95.14               & 0.27                \\
                                     & Long-CoT                     & \textbf{47.56}                  & 97.51                         & \textbf{0.86}       & 63.95               & \textbf{96.58}      & \textbf{0.70}       \\
        \midrule
        \multirow{2}[0]{*}{GPT}      & Short-CoT                    & \textbf{34.34}                  & 71.05                         & 0.26                & 30.79               & 33.52               & 0.16                \\
                                     & Long-CoT                     & 10.41                           & \textbf{92.12}                & \textbf{1.28}       & \textbf{45.05}      & \textbf{99.23}      & \textbf{3.03}       \\
        \bottomrule
    \end{tabular}
    \caption{
        Long-CoT models demonstrate significantly higher rates of Error Correction and New Solution Discovery, indicating a more flexible and reflective reasoning process compared to the cognitive rigidity of Short-CoT.
    }
    \label{tab:stability}
\end{table*}

\subsection{Existence Verification}
\paragraph{Short-CoT displays substantial reasoning overconfidence across all model series.}
To assess the existence of ROC, we quantify this phenomenon using recall–confidence reliability diagrams, in which a perfectly calibrated model aligns with the main diagonal. In Figure~\ref{fig:reliability-diagrams-and-short-cot-long-cot-recall} (a), the Short-CoT bars aggregate in the lower-right quadrant, indicating high confidence yet low recall, and therefore fall well below the diagonal. Consistently, Short-CoT produces expected calibration error (ECE) values above 78.22\% for every model series, corroborating its marked reasoning overconfidence.

\vspace{-4mm}
\paragraph{Long-CoT lowers reasoning over-confidence relative to Short-CoT, yet open-source models still need further improvement.}
Figure \ref{fig:reliability-diagrams-and-short-cot-long-cot-recall} (a) shows that the Long-CoT reliability curve lies closer to the diagonal compared to Short-CoT, indicating better calibration and a lower rate of ROC. Quantitatively, Long-CoT decreases the ECE by at least 18.28\% across all model families and even delivers single-digit ECE for closed-source models. Nevertheless, open-source models continue to exhibit pronounced ROC, exceeding 56.46\%, even after Long-CoT prompting. In sum, although Long-CoT markedly mitigates ROC compared with Short-CoT, additional advances are required to enhance calibration in open-source models.

\subsection{Behavioral Diagnostics}

We demonstrate the cause of ROC under the Short-CoT paradigm. We discover that models meeting with ROC can trigger the following unexpected behavior:

\vspace{-4mm}
\paragraph{Calibrating ROC is coupled with actual performance improvement.}
Short-CoT instances concentrate in the \textbf{“low-recall, high-confidence”} quadrant (Figure \ref{fig:recall-vs-confidence-3d}). By halting early, as shown in Figure \ref{fig:reliability-diagrams-and-short-cot-long-cot-recall} (b), the model retrieves few correct solutions yet remains overly sure of their completeness, constraining answer diversity. Figure \ref{fig:recall-vs-confidence-movement} plots, for each problem, the vector from the Short-CoT point to its Long-CoT counterpart. The consistent upward shift indicates that Long-CoT searches more exhaustively and recovers solutions missed by Short-CoT, while the concurrent leftward shift toward lower confidence reveals better self-calibration. Thus, ROC calibration aligns with performance gains.

\begin{figure*}[t]
    \centering
    \includegraphics[width=0.98\linewidth]{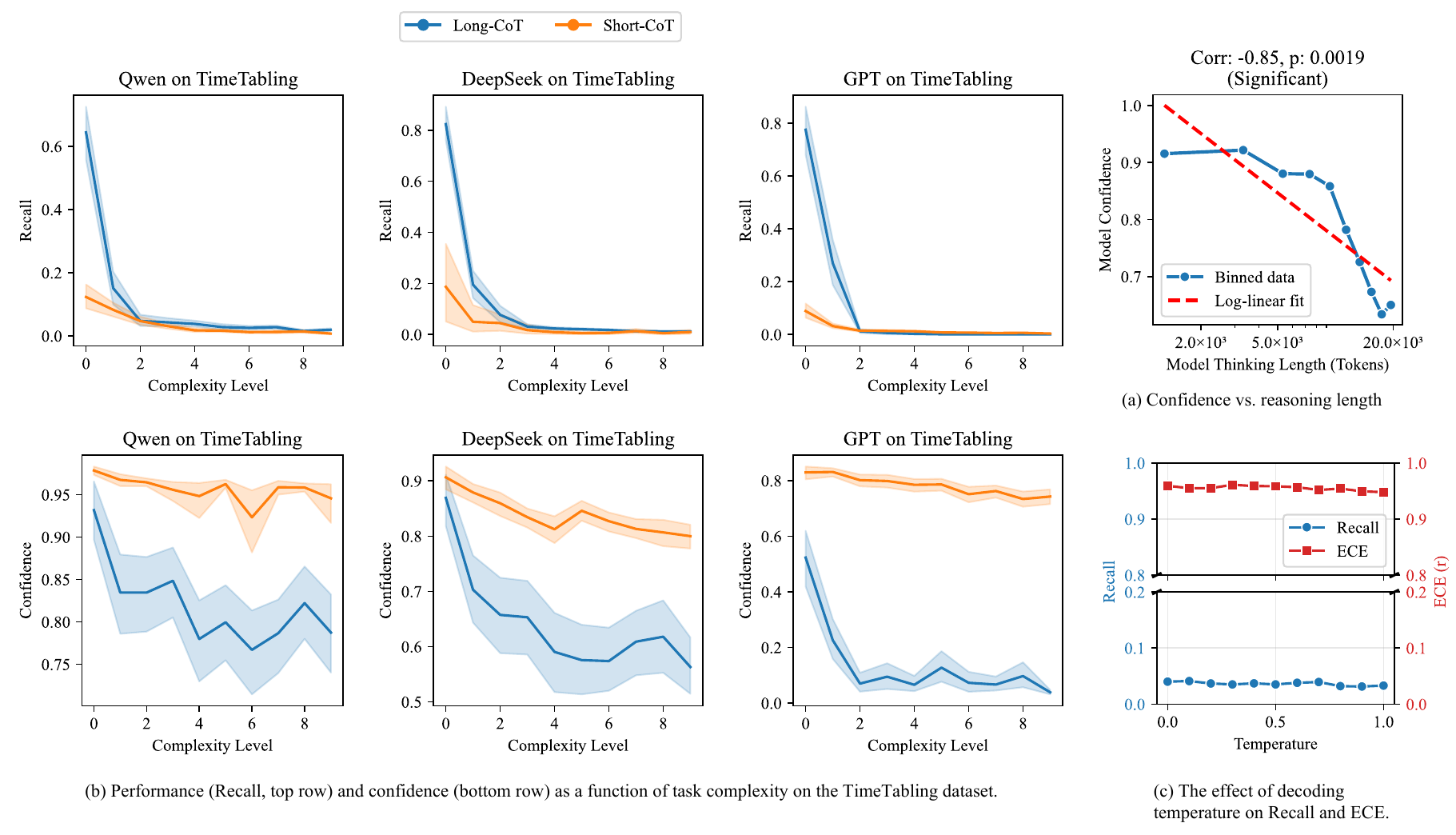}
    \caption{Factors that influence reasoning overconfidence.
        (a) A strong negative correlation shows that Long-CoT has moderate confidence.
        (b) As task complexity rises, Short-CoT keeps unjustifiably high confidence despite falling recall, indicating poor self-monitoring, whereas Long-CoT lowers its confidence in line with the harder setting, demonstrating better calibration.
        (c) Decoding temperature has little effect on recall or expected calibration error.
        More results see in Figure~\ref{fig:complexity-vs-confidence-and-length-and-temperature-subsetsum} in Appendix.}
    \label{fig:complexity-vs-confidence-and-length-and-temperature}
\end{figure*}

\vspace{-4mm}
\paragraph{Cognitive Rigidity and Resistance to Guidance.}
As shown in Table~\ref{tab:stability}, models using Short-CoT exhibit extremely low error correction and new solution discovery rates.
Even when prompted to reconsider, the model is largely unable to identify its previous errors or explore new, correct paths.
This cognitive rigidity indicates that the model stops exploring alternative reasoning paths once it settles on an initial set of answers.
Long-CoT models, in contrast, are far more capable of self-correction and discovery.
Experiments on dataset generalizability are available in Appendix~\ref{appsec:dataset-generalizability}.

\subsection{Influencing Factors}

We now investigate key factors that mitigate ROC behavior.

\vspace{-4mm}
\paragraph{Inference-Time Scaling Law also holds for the phenomenon of reasoning overconfidence.}
Building on the inference-time scaling law, which raises accuracy by increasing inference calculations~\cite{wu2024inference}, we investigate whether longer chains also temper ROC. We therefore measure how reasoning length relates to the model’s final confidence. Figure~\ref{fig:complexity-vs-confidence-and-length-and-temperature} (a) shows a strong negative correlation: extended chains yield lower confidence estimates. Thus, extra computation not only improves accuracy but also calibrates predictions by curbing ROC; longer reasoning promotes a more cautious self-assessment.

\vspace{-4mm}
\paragraph{Long-CoT strategies substantially reduce ROC as task complexity increases, whereas Short-CoT is insensitive to task complexity.}
Following previous work~\cite{xu2025language}, we posit that greater complexity, reflected by a much smaller ground-truth solution space, should dampen a model’s confidence. Figure~\ref{fig:complexity-vs-confidence-and-length-and-temperature}(b) confirms this expectation for Long-CoT: its confidence decreases as complexity rises, indicating proper calibration. Short-CoT, however, maintains high confidence while recall drops sharply, revealing persistent overconfidence. Hence, Long-CoT acknowledges growing difficulty, whereas Short-CoT remains blind to task complexity.

\begin{figure*}[t]
    \centering
    \includegraphics[width=0.98\linewidth]{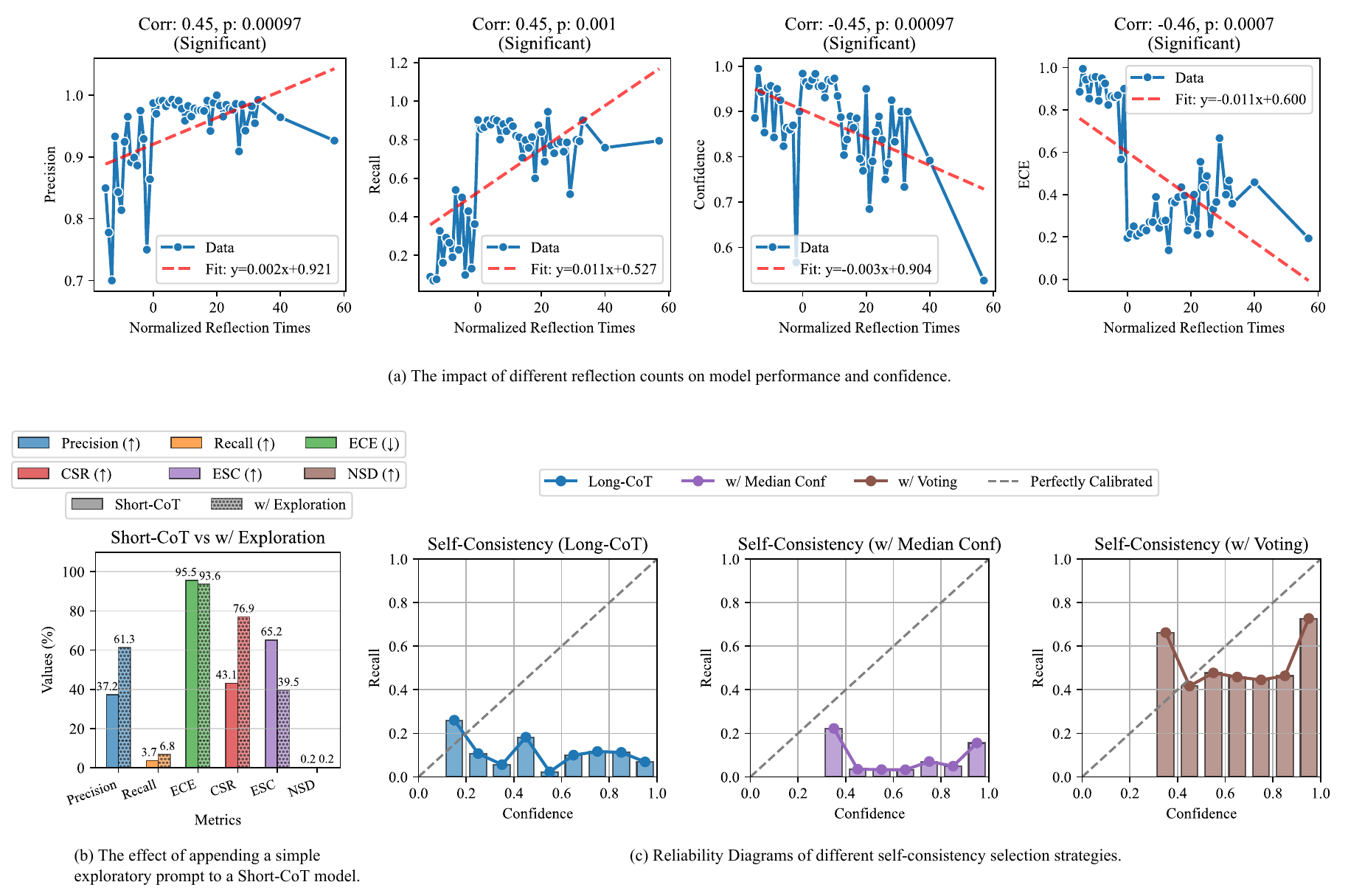}
    \caption{
        Mitigate reasoning overconfidence through various strategies.
        (a)
        As reflection time increases, recall improves, and overconfidence decreases significantly.
        (b)
        A simple exploratory prompt boosts both precision and recall, demonstrating its potential at breaking cognitive rigidity.
        (c)
        \textit{w/ Voting} aggregation strategy significantly improves calibration.
        More results are shown in Figure~\ref{fig:reflection-and-exploration-and-self-consistency-subsetsum} in Appendix.
    }
    \label{fig:reflection-and-exploration-and-self-consistency}
\end{figure*}

\begin{table}[t]
    \centering
    \small
        \begin{tabular}{ccccccc}
        \toprule
        Method         & Precision (\%) $\uparrow$ & Recall (\%) $\uparrow$ & ECE (r) ($\downarrow$) \\
        \midrule
        \rowcolor{gray!20}\multicolumn{4}{c}{TimeTabling}                                            \\
        \midrule
        Long-CoT       & 54.88                     & 9.99                   & 77.23                  \\
        w/ Median Conf & 59.10                     & 19.81                  & 81.13                  \\
        w/ Voting      & \textbf{74.04}            & \textbf{49.55}         & \textbf{64.32}         \\
        \midrule
        \rowcolor{gray!20}\multicolumn{4}{c}{SubsetSum}                                              \\
        \midrule
        Long-CoT       & 85.98                     & 21.24                  & 68.31                  \\
        w/ Median Conf & \textbf{92.13}            & 33.33                  & 71.68                  \\
        w/ Voting      & 72.75                     & \textbf{51.40}         & \textbf{65.43}         \\
        \bottomrule
    \end{tabular}
    \caption{
        Performance of self-consistency strategies on the Long-CoT Qwen3-8B model, using n=32 parallel reasoning paths.
        Results highlight that the choice of aggregation strategy involves significant trade-offs between performance and model calibration.
    }
    \label{tab:self-consistency}
\end{table}

\vspace{-4mm}
\paragraph{Decoding temperature increases token-level diversity rather than alleviating ROC or expanding solution exploration.}
Figure \ref{fig:complexity-vs-confidence-and-length-and-temperature}(c) demonstrates that adjusting the decoding temperature scarcely affects the model’s self-reported confidence and fails to reduce ROC. Higher temperatures likewise do not encourage a broader search across alternative reasoning paths. Instead, temperature primarily injects stochastic variation at the token level, enlarging lexical diversity without prompting the model to revisit or deepen its chain of thought. These observations indicate that ROC originates in the model’s intrinsic reasoning mechanism rather than in tunable decoding heuristics.

\section{Mitigation Strategies}

Building on \citet{chen2025towards}, we separate the two mechanisms thought to drive Long-CoT performance, reflection, and exploration, to quantify their individual effects and derive practical guidance.

\vspace{-4mm}
\paragraph{Reflection steps enhance solution diversity and reduce ROC.}
We evaluated the reflection’s effect on ROC by pausing Long-CoT at scheduled checkpoints. After a fixed number of iterations, and before the final answer, we inserted the control token \verb|</think>| to record the intermediate output. Figure \ref{fig:reflection-and-exploration-and-self-consistency} (a) reveals that additional reflection rounds raise recall while lowering confidence, thereby lessening ROC. This evidence indicates that reflection exposes otherwise overlooked solutions and thus decreases ROC.

\vspace{-4mm}
\paragraph{Sequential exploration–scaling prompts unlock rigid thinking and reduce ROC in Short-CoT.}
We tested whether ROC of Short-CoT can be eased by an external exploratory cue. After the model produced its initial answer, we appended the prompt ``Wait, there may be other solutions.'' Figure \ref{fig:reflection-and-exploration-and-self-consistency} (b) shows that this cue markedly improves performance. The added exploration frees the model from cognitive rigidity and yields more correct answers. ROC also drops slightly, as shown in Figure \ref{fig:exploration-calibration} in Appendix.

\vspace{-4mm}
\paragraph{Parallel exploration–scaling prompting mitigates ROC and boosts recall.}
We use a self-consistency paradigm that generates multiple reasoning paths in parallel and aggregates them by two strategies:
(1) \textit{w/ Median Conf}: select the path with the median confidence score;
(2) \textit{w/ Voting}: unite all unique answers and weight their confidence by frequency.
As shown in Table \ref{tab:self-consistency}, aggregation choice is decisive. \textit{w/ Voting} markedly increases recall on both datasets and improves calibration, indicating stronger ROC mitigation and broader solution coverage. By contrast, as shown in Figure \ref{fig:reflection-and-exploration-and-self-consistency} (c),\textit{Median Conf} raises precision and recall over the baseline but degrades calibration, worsening ECE.

\section{Investigating the Internal Mechanism}

To explain the rationale of reasoning overconfidence, we introduce the \textbf{cognitive-rigidity hypothesis}.
It posits that ROC arises when the model’s core reasoning layers lock too early onto a single trajectory, exhibiting cognitive rigidity. Following \citet{cui2025entropy}, we treat attention entropy as a proxy for internal diversity. Accordingly, we compute layer-wise attention entropy for Qwen3-8B under the Short-CoT and the lower-ROC Long-CoT settings.

\begin{figure*}[t]
    \centering
    \includegraphics[width=0.98\linewidth]{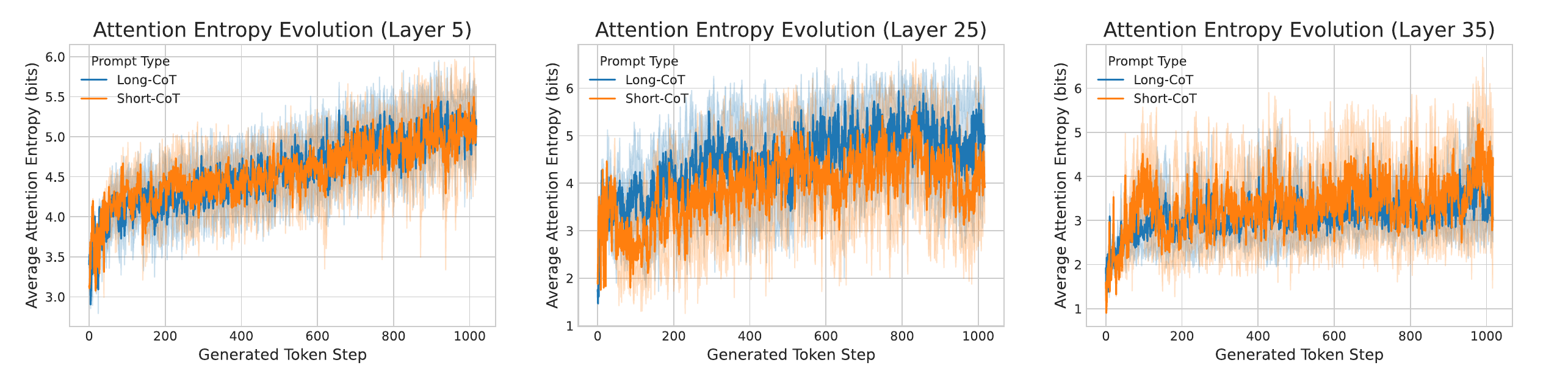}
    \caption{Attention entropy of Short-CoT vs. Long-CoT on Qwen3-8B. We present layers 5, 25, and 35 as representative examples of three phases of the reasoning process. The model’s entropy evolution follows a three-phase pattern of convergence, divergence, and reversal: the Long-CoT paradigm shows higher entropy in core layers, promoting exploration, whereas the Short-CoT paradigm ends with higher entropy in the deepest layers. Detailed results are provided in Figure~\ref{fig:attention-entropy} in the Appendix.}
    \label{fig:mechanism-entropy-layer25}
\end{figure*}

\vspace{-4mm}
\paragraph{Paradigms with reduced ROC show low attention entropy in core reasoning layers, indicating cognitive rigidity.}
Figure \ref{fig:mechanism-entropy-layer25} exhibits three phases consistent with \citet{cui2025entropy}:
(1) Shallow layers (0–10): Initial entropies differ but rapidly converge.
(2) Core reasoning layers (15–30): Reduced-ROC paradigms sustain higher entropy than high-ROC counterparts.
(3) Deep layers ($\approx$35): The trend reverses; high-ROC paradigms terminate with greater entropy.

These patterns support the cognitive-rigidity hypothesis. In core layers, the low entropy of high-ROC paradigms signals narrowly focused attention that restricts exploratory reasoning, whereas Long-CoT’s higher entropy reflects flexibility to pursue alternative paths. The final reversal strengthens this interpretation: high-ROC rigidity produces late-stage uncertainty (high entropy), while Long-CoT, having explored more broadly, converges decisively (low entropy).

\section{Related Work}

Chain-of-Thought (CoT) prompting markedly elevates Large Language Models' reasoning by eliciting explicit intermediate steps that mirror human cognition \cite{wei2022chain,chen2024unlocking,chen2024m,cheng2025visual}. Yet its standard, concise variant, Short-CoT, remains brittle~\citep{qin2023cross,chen2025towards}. To improve robustness, later work introduced more expressive reasoning topologies. Tree-of-Thought \cite{yao2023tree} and Graph-of-Thoughts \cite{besta2024graph}, for example, let models explore multiple reasoning paths concurrently, enabling branching and backtracking to boost the likelihood of a correct answer~\citep{hu2024treeplanner}.

A model's ability to report accurate confidence is crucial for its reliability, especially in high-stakes applications and for detecting hallucinations~\cite{amodei2016concrete,hou2024probabilistic}.
However, a significant body of work has shown that LLMs are often poorly calibrated and exhibit overconfidence, expressing high certainty in answers that are incorrect or incomplete~\cite{jiang2020can,singh2023confidence}.
Research has explored mitigating this issue through methods like probability calibration~\cite{fisch2022calibrated,guo2017calibration} or by developing prompting strategies that leverage self-consistency or relative ranking to better estimate verbal confidence~\cite{li2024confidence,yang2024confidence,shrivastava2025language}.
Overconfidence is pronounced in CoT reasoning: orderly steps can falsely signal correctness, leading models to commit before exploring the full solution space \cite{ling2023deductive, cheng2025chain}. Although recent work shows Long-CoT models are better calibrated than Short-CoT ones \cite{yoon2025reasoning}, these studies, and calibration research in general, still judge confidence only by the accuracy of a single final answer.

Prior efforts that allow multiple solutions either restrict the solution space to narrow fact-checking tasks \cite{lin2021truthfulqa, li2024think} or apply standard calibration metrics \cite{lin2022teaching} without addressing the unique failure modes of exhaustive generation. 
To examine a key failure mode in multi-solution scenarios, we formally define and analyze Reasoning Overconfidence: a model’s unwarranted certainty that its generated solution set is complete. We shift evaluation from confidence-accuracy to confidence-completeness correlations, offering a fresh lens for enhancing LLM reliability.

\section{Conclusion}

This paper identifies reasoning overconfidence in LLMs as a critical failure mode on multi-solution tasks, where standard Short-CoT yields incomplete yet highly confident solutions.
To better understand this phenomenon, our analysis delves into its key influencing factors.
In response to this problem, we demonstrate that the emerging Long-CoT effectively mitigates this issue by improving both solution diversity and confidence calibration.
We attribute the success of Long-CoT to its ability to overcome our proposed cognitive-rigidity hypothesis: a state where Short-CoT locks the model into a narrow search space.
These findings underscore the limitations of conventional CoT and call for more exploratory reasoning paradigms to build reliable AI.

\section*{Acknowledgements}

This work was supported by the National Natural Science Foundation of China (NSFC) via grants 62236004, 62206078, 62476073, 92570120, and 62306342.
This work was sponsored by the CCF-Zhipu Large Model Innovation Fund (NO.CCF-Zhipu202406).

\end{CJK}

\bibliographystyle{./refstyle}
\bibliography{ref}
\appendix
\newpage
\section{Appendix}

\subsection{Benchmark Construction}
\label{appsec:benchmark-construction}

To evaluate reasoning overconfidence on multi-solution tasks, we constructed two unique datasets focused on combinatorial optimization and constraint satisfaction problems: Timetabling and SubsetSum.
The construction method for both datasets follows a systematic three-stage process.
This process includes (1) parameterized problem generation, (2) exhaustive solution enumeration through deterministic search, and (3) complexity stratification based on the size of the solution space.
Using this method, we produced a collection of problems that have \textit{known and complete} solution sets.
These problems are also categorized according to a quantifiable complexity metric.

The first dataset contains instances of the Timetabling problem, which is a classic constraint satisfaction problem.
Each problem instance defines a set of resources and constraints.
The objective is to generate all valid schedules by assigning a specific time slot and room to each course without violating any constraints.
A problem is formally defined by a set of courses, time slots, rooms, and teachers.
The constraints for a valid schedule are as follows:
Each course must be assigned to one of its pre-approved time slots and rooms.
Each course is assigned a unique teacher, and a teacher cannot teach multiple courses at the same time.
Furthermore, a room cannot accommodate more than one course at any given time.

The generation process begins by randomly determining the number of courses, teachers, rooms, and time slots within preset limits.
A teacher is then randomly assigned to each course.
Next, the set of allowed time slots and rooms for each course is generated by randomly sampling from the global pool of available times and rooms.
To introduce variety in constraint tightness, we use a probabilistic parameter.
This parameter controls whether a course is restricted to a single, specific room, which creates more constrained problem instances.

The second dataset contains instances of the SubsetSum problem.
This is a classic NP-complete problem in computational complexity theory.
A problem instance consists of a set of unique integers, $S = \{s_1, s_2, \dots, s_n\}$, and a target integer value $K$.
The task is to find all non-empty subsets $S' \subseteq S$ where the sum of the elements is exactly equal to the target, such that $\sum_{s_i \in S'} s_i = K$.

The generation process for these problems is designed to guarantee that each instance has at least one solution.
This approach avoids trivial cases where no solution exists.
First, a set $S$ containing $n$ unique integers is created by sampling from a specified value range.
Subsequently, instead of choosing a random target $K$ that might not be achievable, the target is constructed differently.
A random, non-empty subset of $S$ is selected, and the sum of its elements is calculated.
This sum is then set as the problem's target $K$.
This construction method ensures that at least one feasible subset exists.

For both datasets, once a problem instance is generated, a deterministic backtracking algorithm is used to systematically explore the entire search space.
This exhaustive search finds every possible valid solution, whether it is a complete schedule or a qualifying subset.
The total number of solutions is then counted.
This count serves as a practical indicator of the problem's combinatorial complexity.
Problems with a larger solution space are considered to be in a different complexity class than those with a smaller one.

Based on this solution count, each problem is assigned to a specific complexity level.
These levels are defined by ranges in the number of solutions, for example, 1-50 solutions or 51-100 solutions.
The generation and solving loop continues until a target quota of problems is met for each complexity level.
Finally, each generated problem and its complete, enumerated set of solutions are formatted into human-readable text.
They are then provided as question-answer pairs for manual verification.

Prompt for Long-CoT models on TimeTabling task:
\begin{promptbox}

    \textbf{[Instruction$_{1,TimeTabling}$]}
    \begin{verbatim}
You are asked to perform a timetabling task.
Please find ALL FEASIBLE SCHEDULES that satisfies all constraints one by
one and output the number of feasible schedules.
Output format example:
Solution 1:
| Course  | Time  | Room  | Teacher  |
|---------|-------|-------|----------|
| Course0 | T2    | R0    | P0       |
| Course1 | T3    | R2    | P2       |
| Course2 | T0    | R2    | P1       |
Solution 2:
| Course  | Time  | Room  | Teacher  |
|---------|-------|-------|----------|
| Course0 | T2    | R0    | P0       |
| Course1 | T3    | R2    | P2       |
| Course2 | T1    | R2    | P1       |
...

Total xxx feasible solutions shown above.

The question is <<QUESTION>>
You must output all feasible solutions without using ellipsis, etc.
The most important thing is to FIND THE SPECIFIC CONTENT OF EACH SOLUTION,
rather than just counting the number of solutions.
Please note that the examples I gave you are just to show the format, the
actual answer may be different from the examples shown.
    \end{verbatim}
\end{promptbox}
where \verb|<<QUESTION>>| is a task instance of the TimeTabling task.

Prompt for Long-CoT models on SubsetSum task:
\begin{promptbox}

    \textbf{[Instruction$_{1,SubsetSum}$]}
    \begin{verbatim}
You are asked to perform a subset-sum 
task.
Please find ALL FEASIBLE SUBSETS that meet the requirements one by 
one and output the number of feasible subsets.
Output format example:
Solution 1: {1, 3, 5}
Solution 2: {1, 4, 4}
...

Total xxx feasible solutions shown above.

The question is <<QUESTION>>
You must output all feasible solutions without using ellipsis, etc.
    \end{verbatim}
\end{promptbox}

\begin{promptbox}

    \begin{verbatim}
The most important thing is to FIND THE SPECIFIC CONTENT OF EACH SOLUTION,
rather than just counting the number of solutions.
Please note that the examples I gave you are just to show the format, the
actual answer may be different from the examples shown.
    \end{verbatim}
\end{promptbox}
where \verb|<<QUESTION>>| is a task instance of the SubsetSum task.

For Short-CoT models, we append a guiding phrase ``Think step by step before answering.'' to the instruction.

\begin{figure*}[h]
    \centering
    \includegraphics[width=0.95\linewidth]{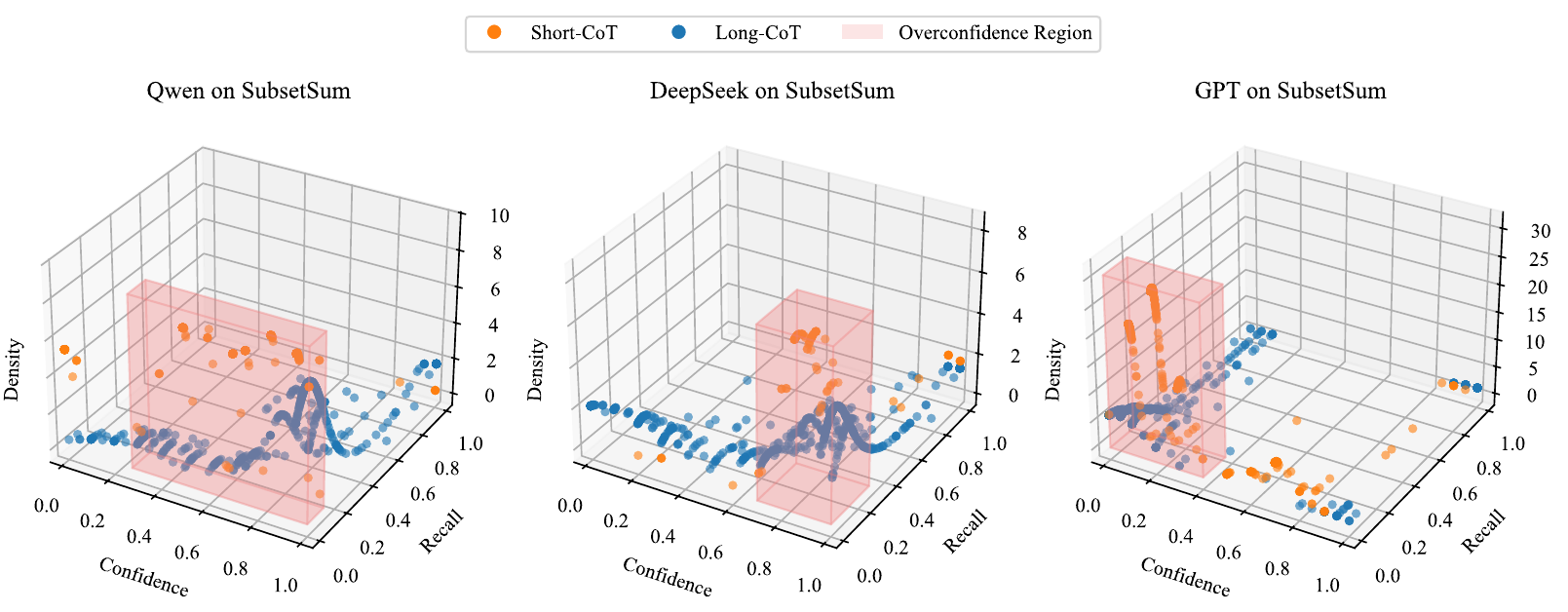}
    \caption{
        Distribution plots of recall vs. confidence on the SubsetSum dataset.
    }
    \label{fig:recall-vs-confidence-3d-subsetsum}
\end{figure*}

\begin{figure*}[h]
    \centering
    \includegraphics[width=0.95\linewidth]{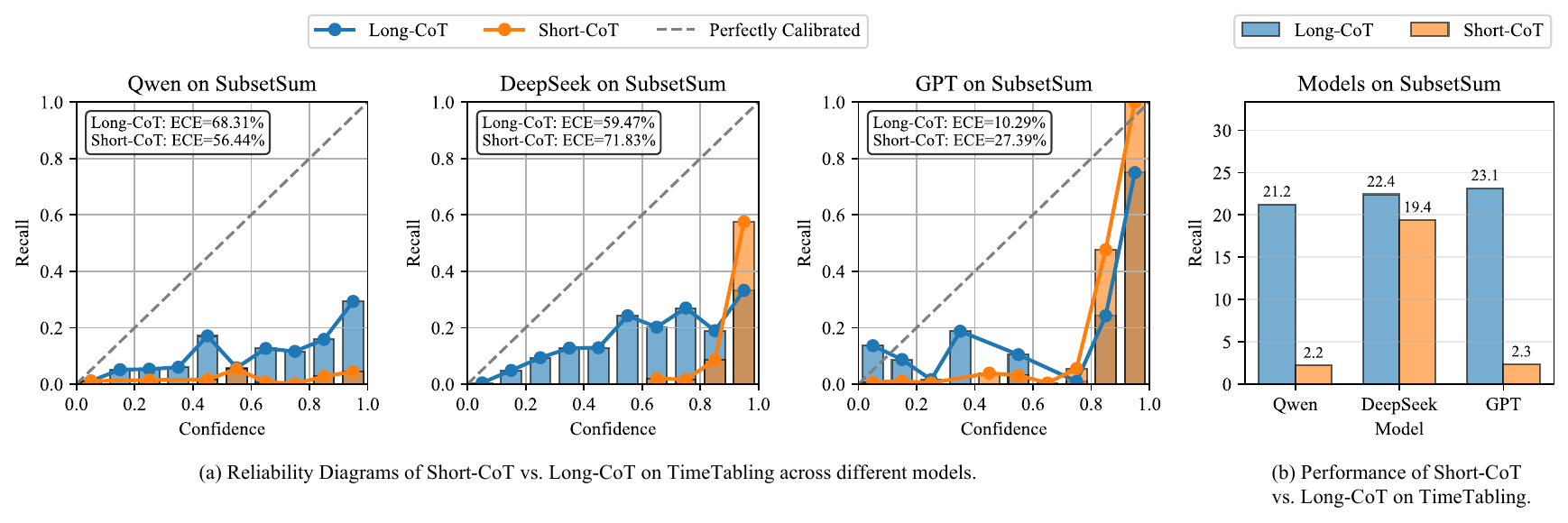}
    \caption{
        Calibration and performance of Short-CoT vs. Long-CoT on SubsetSum dataset.
    }
    \label{fig:reliability-diagrams-and-short-cot-long-cot-recall-subsetsum}
\end{figure*}

Example of TimeTabling task:
\begin{promptbox}

    \textbf{[Question]}
    \begin{verbatim}
Constraints:
- Course0 : Time [3, 4], Room [8], 
Teacher [2]
- Course1 : Time [0, 4], Room [3], 
Teacher [2]
- Course2 : Time [3, 4], Room [0], 
Teacher [1]
- Multiple courses cannot be scheduled in the same time slot and room.
- A teacher can only teach one course at a time.
    \end{verbatim}
    \textbf{[Answer]}
    \begin{verbatim}
Solution 1:
| Course  | Time  | Room  | Teacher  |
|---------|-------|-------|----------|
| Course0 | T3    | R8    | P2       |
| Course1 | T0    | R3    | P2       |
| Course2 | T3    | R0    | P1       |
Solution 2:
| Course  | Time  | Room  | Teacher  |
|---------|-------|-------|----------|
| Course0 | T3    | R8    | P2       |
| Course1 | T0    | R3    | P2       |
| Course2 | T4    | R0    | P1       |
Solution 3:
| Course  | Time  | Room  | Teacher  |
|---------|-------|-------|----------|
| Course0 | T3    | R8    | P2       |
| Course1 | T4    | R3    | P2       |
| Course2 | T4    | R0    | P1       |
Solution 4:
...
    \end{verbatim}
\end{promptbox}

Example of SubsetSum task:
\begin{promptbox}

    \textbf{[Question]}
    \begin{verbatim}
Given the set of unique integers: 
{18, 25, 16, 45, 48, 40, 38, 14, 22, 8}
Find all subsets that sum exactly to the target: 62
    \end{verbatim}
    \textbf{[Answer]}
    \begin{verbatim}
[8, 14, 18, 22], [8, 14, 40], [8, 16, 38], [14, 48], [22, 40]
    \end{verbatim}
\end{promptbox}

\subsection{Evaluation Metric Construction}
\label{appsec:evaluation-metrics}

We use the following metrics to evaluate model behavior quantitatively:

\textbf{Precision:}
The fraction of generated answers that are correct.
\begin{equation}
    \text{Precision}(x) = \frac{|\mathcal{Y}_i \cap \hat{\mathcal{Y}}_i|}{|\mathcal{Y}_i|}
\end{equation}

\textbf{Recall:}
The fraction of ground-truth answers that the model found.
\begin{equation}
    \text{Recall}(x) = \frac{|\mathcal{Y}_i \cap \hat{\mathcal{Y}}_i|}{|\hat{\mathcal{Y}}_i|}
\end{equation}
This is our primary metric for multi-solution tasks.

\textbf{Expected Calibration Error (ECE):}
ECE measures the consistency between a model's reported confidence and its average performance within specific confidence intervals.
A perfectly calibrated model has an ECE of 0.

The calculation process is as follows:
First, the confidence interval $[0, 1]$ is divided into $M$ equally-sized bins, denoted as $B_1, B_2, \dots, B_M$.
For all $N$ problem instances, each is assigned to its corresponding bin based on the model's reported confidence score.
For each bin $B_m$, the average confidence $\text{conf}(B_m)$ and the average performance (e.g., average precision $\text{prec}(B_m)$ or average recall $\text{recall}(B_m)$) of the samples within it are calculated.
\begin{align}
    \text{conf}(B_m)   & = \frac{1}{|B_m|} \sum_{i \in B_m} c_i                   \\
    \text{prec}(B_m)   & = \frac{1}{|B_m|} \sum_{i \in B_m} \text{Precision}(x_i) \\
    \text{recall}(B_m) & = \frac{1}{|B_m|} \sum_{i \in B_m} \text{Recall}(x_i)
\end{align}
Here, $|B_m|$ is the number of samples in bin $m$, $c_i$ is the confidence of the $i$-th sample, and $\text{Precision}(x_i)$ and $\text{Recall}(x_i)$ are its precision and recall, respectively.

\begin{figure*}[t]
    \centering
    \includegraphics[width=0.9\linewidth]{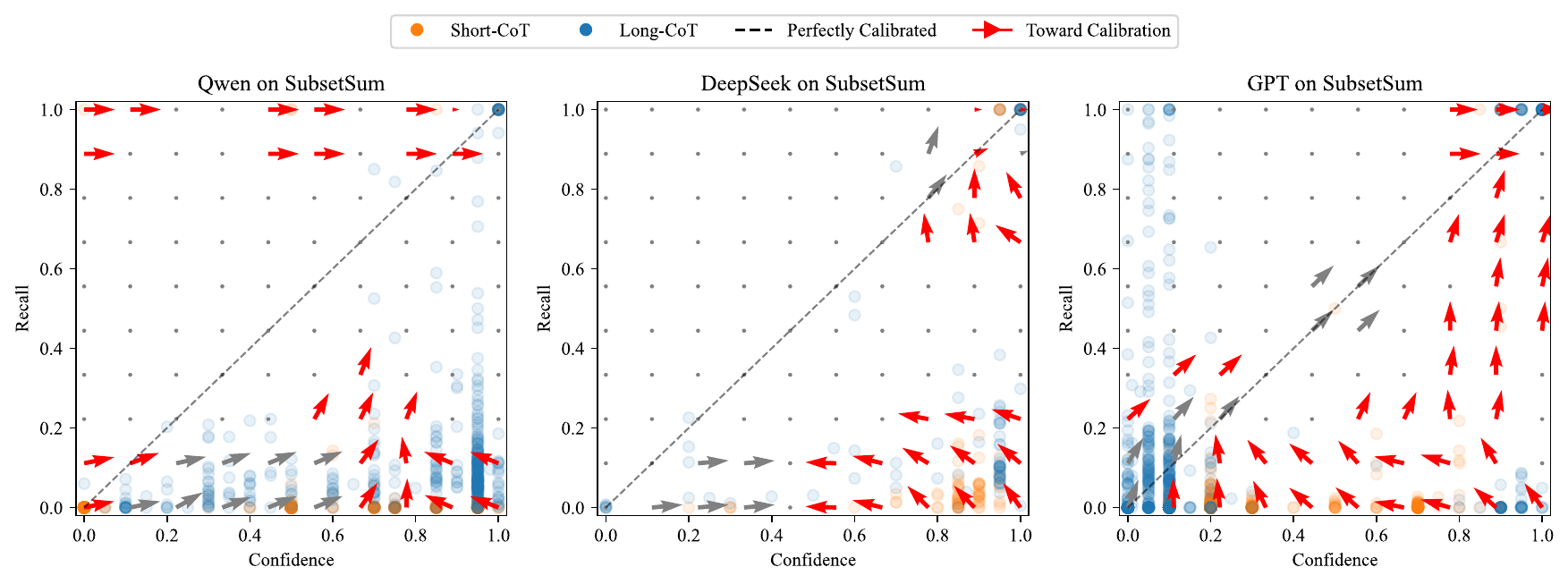}
    \caption{
        The arrows indicate the movement of model confidence and performance from Short-CoT to Long-CoT.
    }
    \label{fig:recall-vs-confidence-movement-subsetsum}
\end{figure*}

The ECE value is the weighted average of the absolute difference between the average performance and the average confidence across all bins.
Depending on the performance metric, we define ECE based on precision ($\text{ECE (p)}$) and ECE based on recall ($\text{ECE (r)}$).
\begin{align}
    \text{ECE (p)} & = \sum_{m=1}^{M} \frac{|B_m|}{N} |\text{prec}(B_m) - \text{conf}(B_m)|   \\
    \text{ECE (r)} & = \sum_{m=1}^{M} \frac{|B_m|}{N} |\text{recall}(B_m) - \text{conf}(B_m)|
\end{align}

A lower ECE value indicates better calibration performance, meaning the model's reported confidence is more reliable.

To measure reasoning stability, we ask the model a follow-up question to probe its ability to reflect.
Let $\mathcal{Y}_{i,1}$ and $\mathcal{Y}_{i,2}$ be the solution sets from the first and second rounds.

\textbf{Correct Solution Retention Rate:}
The model's ability to stick to its correct answers.
\begin{equation}
    \text{Retention}(x) = \frac{|(\mathcal{Y}_{i,1} \cap \hat{\mathcal{Y}}_i) \cap (\mathcal{Y}_{i,2} \cap \hat{\mathcal{Y}}_i)|}{|(\mathcal{Y}_{i,1} \cap \hat{\mathcal{Y}}_i)|}
\end{equation}

\textbf{Error Correction Rate:}
The model's ability to fix its mistakes.
\begin{equation}
    \text{Correction}(x) = 1 - \frac{|(\mathcal{Y}_{i,1} \setminus \hat{\mathcal{Y}}_i) \cap (\mathcal{Y}_{i,2} \setminus \hat{\mathcal{Y}}_i)|}{|(\mathcal{Y}_{i,1} \setminus \hat{\mathcal{Y}}_i)|}
\end{equation}

\textbf{New Solution Discovery Rate:}
The model's ability to find new correct solutions.
\begin{equation}
    \text{Discovery}(x) = \frac{|(\mathcal{Y}_{i,2} \cap \hat{\mathcal{Y}}_i) \setminus \mathcal{Y}_{i,1}|}{|\hat{\mathcal{Y}}_i|}
\end{equation}

\subsection{Confidence Elicitation}
\label{appsec:confidence-elicitation}

Prompt for asking model confidence:
\begin{promptbox}

    \textbf{[Instruction$_2$]}
    \begin{verbatim}
Please rate your confidence in the proposed answer on a scale of 0-100.
Put your confidence score within [[CONFIDENCE: \boxed{}]]
    \end{verbatim}
\end{promptbox}

Prompt for model reconsideration:
\begin{promptbox}

    \textbf{[Instruction$_3$]}
    \begin{verbatim}
Recheck all your answers. You can now supplement and correct your answers.
If you think your answer does not need to be changed, please output 
[[UNCHANGE]].
If you need to supplement or correct your answer, please input [[CHANGE]] 
and re-output your new answer IN FULL, NOT JUST THE PART YOU CHANGED.
    \end{verbatim}
\end{promptbox}

\begin{figure*}[h]
    \centering
    \includegraphics[width=0.95\linewidth]{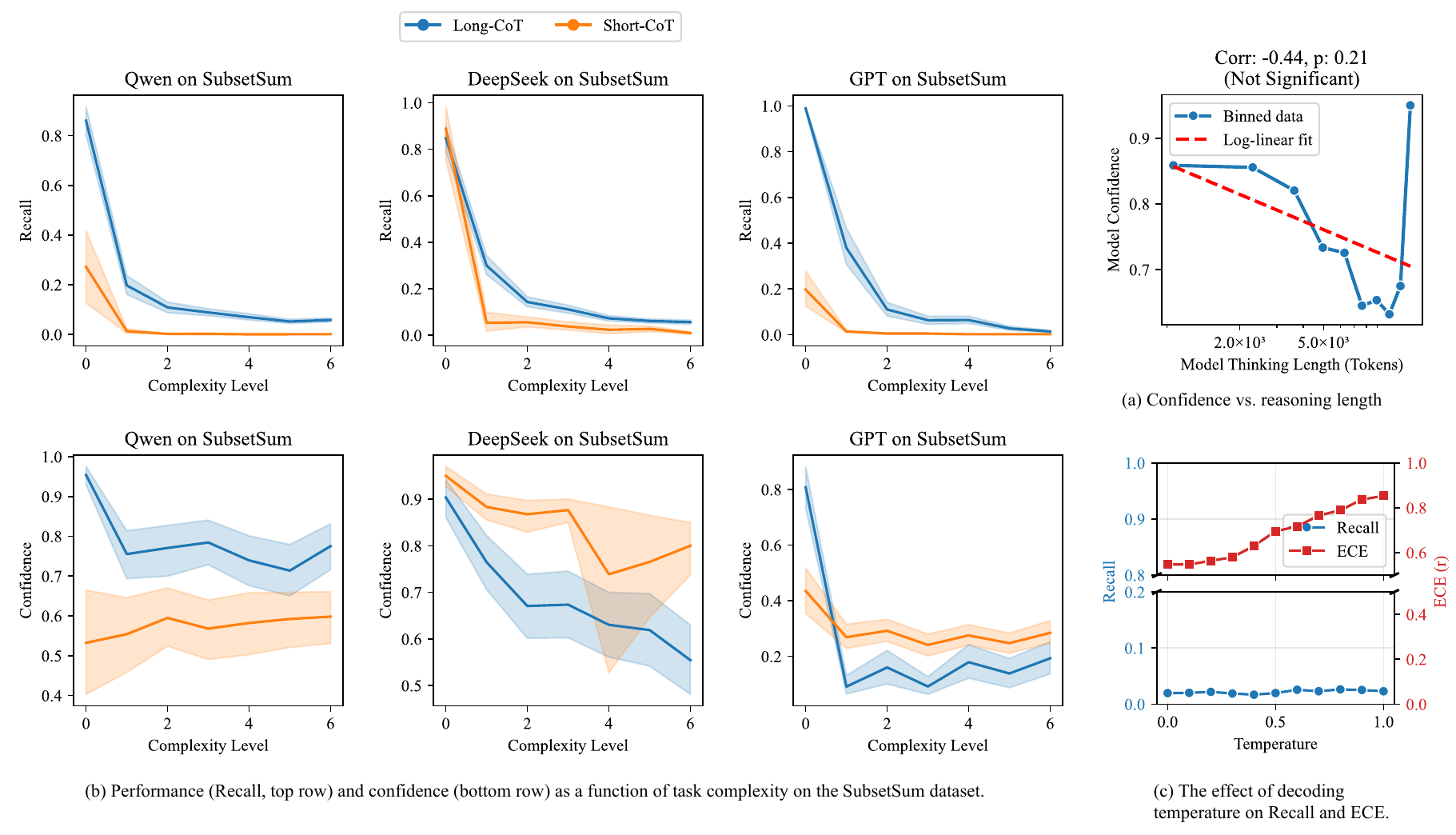}
    \caption{
        Factors that influence reasoning overconfidence.
    }
    \label{fig:complexity-vs-confidence-and-length-and-temperature-subsetsum}
\end{figure*}

\begin{table}[h]
    \centering
    \small
    \begin{tabular}{llr}
        \toprule
        Hyperparameter               & Value         \\
        \midrule
        \verb|temperature|           & 0.2           \\
        \verb|top_p|                 & 1.0 (Default) \\
        \verb|top_k|                 & NA            \\
        \verb|max_completion_tokens| & 20480         \\
        \bottomrule
    \end{tabular}
    
    \caption{
        Hyperparameters used for all experiments.
        Temperature is set to 0.2 except where noted otherwise.
    }
    \label{tab:hyperparameters}
\end{table}

\subsection{Dataset Generalizability}
\label{appsec:dataset-generalizability}

To verify that ROC is not limited to combinatorial tasks, we conducted experiments on NoveltyBench~\cite{zhang2025noveltybench}, a creative generation benchmark.
As shown in Table~\ref{tab:noveltybench}, the results are consistent:
Short-CoT exhibits higher ECE, lower Utility, and lower Distinct compared to Long-CoT, confirming that reasoning overconfidence is a general phenomenon in multi-solution tasks.

\begin{table*}[t]
    \centering
    \small
    \begin{tabular}{cccccc}
        \toprule
        Model                    & Paradigm  & Utility (\%) $\uparrow$ & ECE(Utility) (\%) $\downarrow$ & Distinct (\%) $\uparrow$ & ECE(Distinct) (\%) $\downarrow$ \\
        \midrule
        \multirow{2}[0]{*}{Qwen} & Short-CoT & 41.12                   & 53.87                          & 36.80                    & 65.73                           \\
                                 & Long-CoT  & \textbf{48.34}          & \textbf{47.33}                 & \textbf{45.20}           & \textbf{56.13}                  \\
        \bottomrule
    \end{tabular}
    \caption{
        Results on NoveltyBench.
        Long-CoT consistently outperforms Short-CoT in both performance (Utility and Distinct) and calibration (ECE).
    }
    \label{tab:noveltybench}
\end{table*}

\subsection{Detailed Results}

\begin{figure*}[t]
    \centering
    \includegraphics[width=0.95\linewidth]{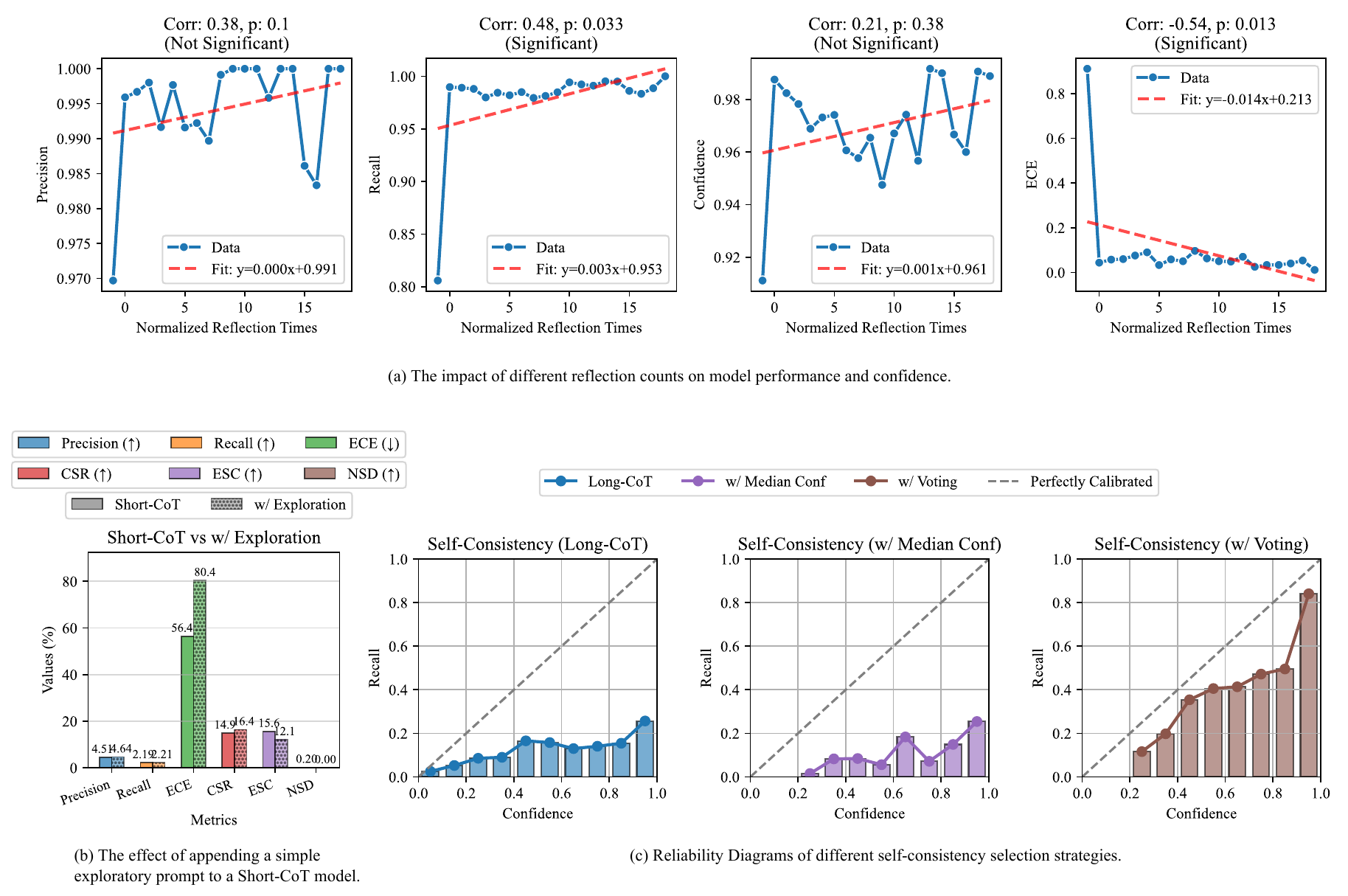}
    \caption{
        Mitigate reasoning overconfidence through various strategies.
    }
    \label{fig:reflection-and-exploration-and-self-consistency-subsetsum}
\end{figure*}

\begin{figure*}[t]
    \centering
    \includegraphics[width=0.95\linewidth]{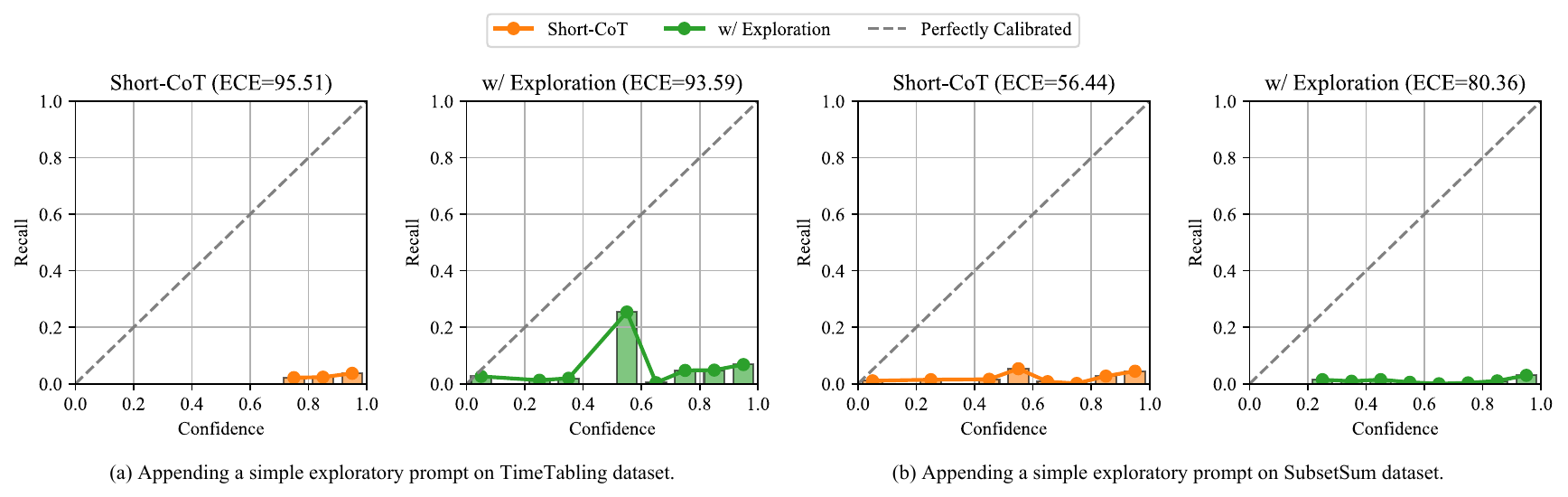}
    \caption{
        Simple exploratory prompt improves calibration on TimeTabling dataset, but not on SubsetSum dataset.
    }
    \label{fig:exploration-calibration}
\end{figure*}

\begin{figure*}[t]
    \centering
    \includegraphics[width=0.95\linewidth]{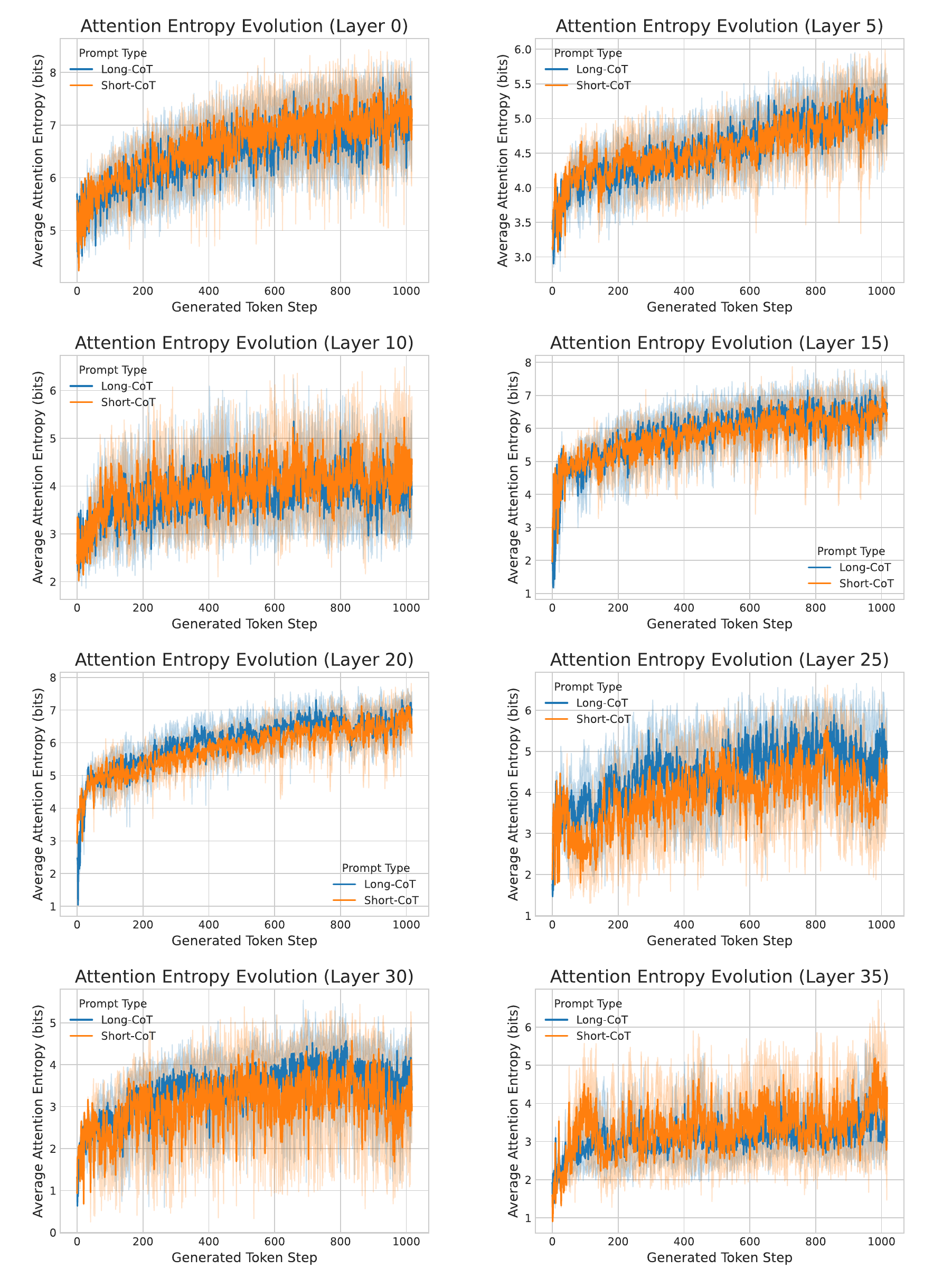}
    \caption{Attention Entropy.}
    \label{fig:attention-entropy}
\end{figure*}

Results on SubsetSum are provided in Figure~\ref{fig:recall-vs-confidence-3d-subsetsum}-\ref{fig:reflection-and-exploration-and-self-consistency-subsetsum}.
Reliability diagrams of the simple exploration prompt are provided in Figure~\ref{fig:exploration-calibration}.
Attention entropy of Qwen3-8B layer 5-35 is shown in Figure~\ref{fig:attention-entropy}.

\subsection{Hyperparameters}

Hyperparameters used in all experiments are shown in Table~\ref{tab:hyperparameters}.
We use OpenAI's API client to access each model, so \verb|top_k| parameter is not applicable.

\subsection{Infrastructure}

All experiments were conducted on an A800 cluster with CUDA driver version 12.4.

\end{document}